\documentclass[journal,comsoc]{IEEEtran}

\usepackage[T1]{fontenc}

\ifCLASSINFOpdf
\else
\fi

\usepackage{amsmath}
\usepackage{verbatim}
\usepackage{bm}
\usepackage{graphicx}
\usepackage{enumerate}
\usepackage{amsfonts}
\usepackage{subfigure}
\usepackage{bbding}
\usepackage{cite}
\usepackage{color}
\usepackage{amsbsy}
\usepackage{amsmath}
\usepackage{indentfirst}
\usepackage{booktabs}

\usepackage{algorithm}
\usepackage{algorithmic}
\usepackage{newfloat}
\usepackage{listings}
\usepackage{ulem}
\floatstyle{ruled}
\newfloat{listing}{tb}{lst}{}
\floatname{listing}{Listing}

\usepackage{latexsym}
\usepackage{amssymb}
\usepackage{amsmath}
\usepackage{adjustbox}
\usepackage{subfigure}

\usepackage{microtype}
\usepackage{multirow}
\usepackage{xcolor}
\usepackage{array}

\usepackage[cmintegrals]{newtxmath}

\begin{document}

\title{Locate before Answering: Answer Guided Question Localization for Video Question Answering}

\author{Tianwen~Qian,
        Ran~Cui,
        Jingjing~Chen,
        Pai~Peng,
        Xiaowei~Guo,
        and~Yu-Gang~Jiang
\thanks{This project was supported by National Key R\&D Program of China (No. 2020AAA0140001), Science and Technology Commission of Shanghai Municipality (No. 21JC1400600). Jingjing Chen is the corresponding author.\\
Tianwen Qian, Jingjing Chen and Yu-Gang Jiang are with Fudan University, Shanghai, China (e-mail: \{twqian19, chenjingjing, ygj\}@fudan.edu.cn). Ran Cui is with Australian National University, Canberra, Australia (e-mail: ran.cui@anu.edu.au). Pai Peng and Xiaowei Guo are with bilibili company, Shanghai, China (e-mail:\{pengpai, weide\}@bilibili.com).}
}

\markboth{IEEE TRANSACTIONS ON MULTIMEDIA}%
{Shell \MakeLowercase{\textit{et al.}}: Bare Demo of IEEEtran.cls for IEEE Communications Society Journals}

\maketitle

\begin{abstract}
Video question answering (VideoQA) is an essential task in vision-language understanding, which has attracted numerous research attention recently. Nevertheless, existing works mostly achieve promising performances on short videos of duration within 15 seconds. For VideoQA on minute-level long-term videos, those methods are likely to fail because of lacking the ability to deal with noise and redundancy caused by scene changes and multiple actions in the video. Considering the fact that the question often remains concentrated in a short temporal range, we propose to first locate the question to a segment in the video and then infer the answer using the located segment only. Under this scheme, we propose ``Locate before Answering'' (LocAns), a novel approach that integrates a question localization module and an answer prediction module into an end-to-end model. During the training phase, the available answer label not only serves as the supervision signal of the answer prediction module, but also is used to generate pseudo temporal labels for the question localization module. Moreover, we design a decoupled alternative training strategy to update the two modules separately. In the experiments,
LocAns achieves state-of-the-art performance on three modern long-term VideoQA datasets, NExT-QA, ActivityNet-QA, and AGQA. Its qualitative examples show the reliable performance of the question localization.
\end{abstract}


\IEEEpeerreviewmaketitle

\section{Introduction}

\IEEEPARstart{V}{ideo} Question Answering (VideoQA) \cite{jang2017tgif, zhu2017uncovering, zhang2019frame} aims to answer a free-form question in natural language based on the contents of a video. As an important task in vision-language understanding, VideoQA is increasingly gathering research attention in recent years \cite{gao2018motion, patel2021recent, khurana2021video} for its potential in applications such as video-based education and robotics.

\begin{figure}[t]
    \centering
    \includegraphics[width=\linewidth]{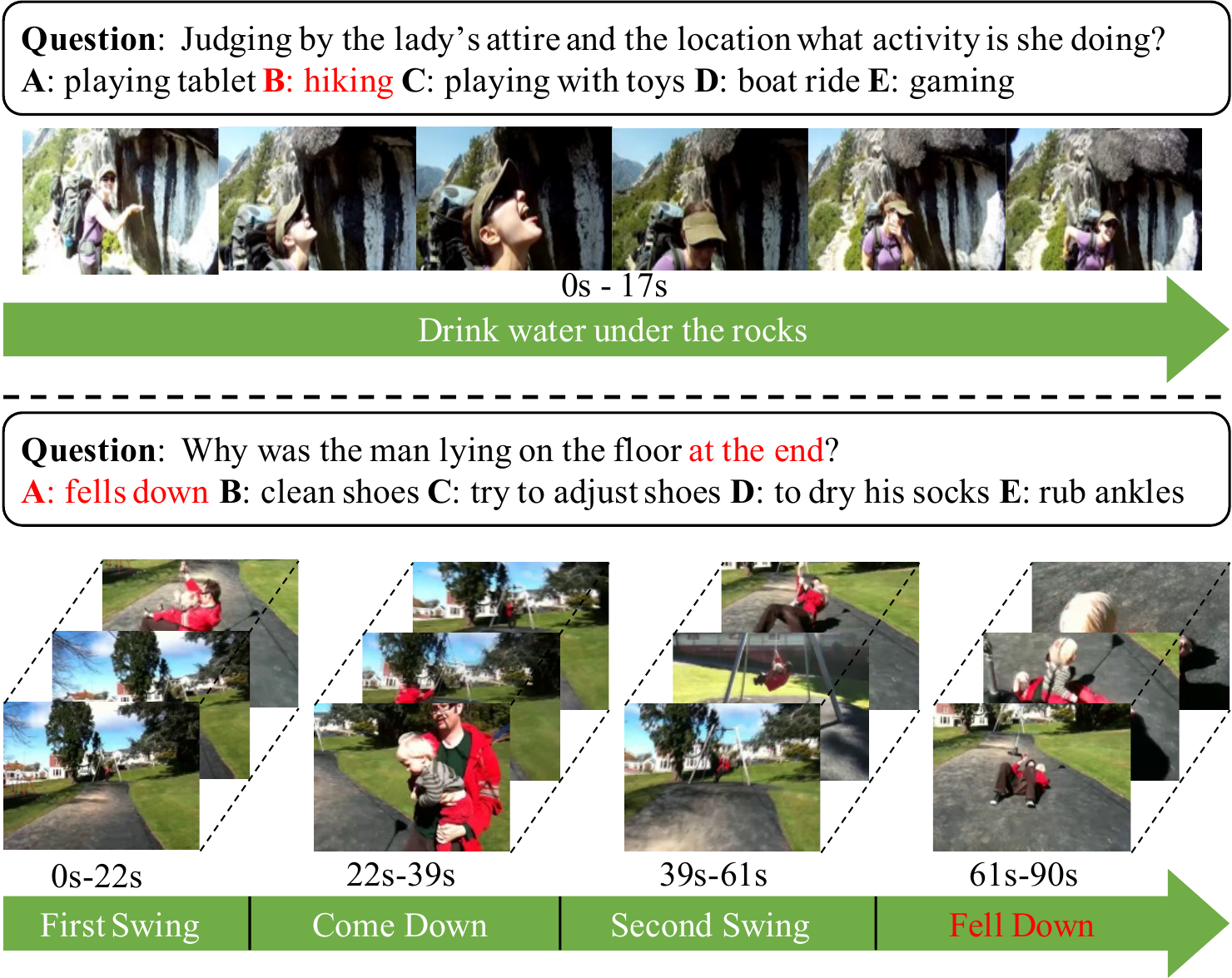}
    \caption{Two examples from the NExT-QA dataset. As the duration gets longer, the video semantics becomes more complex. Hence, it is more likely that the question targets on a small part of the long-term video.}
    \label{fig:introduction}
\end{figure}

Differing to image-based Visual Question Answering (VQA) \cite{malinowski2014multi, antol2015vqa, qian2022scene}, a core problem of VideoQA is how to properly process the visual sequential relation along the time dimension. In the early stage of VideoQA research, most studies \cite{le2020hierarchical, jiang2020divide, wang2021dualvgr} were on short videos within 15 seconds. The limited video duration results in the visual semantics being relatively monotonic, reflected in unchanged scene and consistent action. Thus, those methods tend to treat the video as a whole and encode the overall semantic of the video (such as extracting video-level feature by pooling over all visual tokens in the time dimension). However, long-term videos (minute-level duration) tend to contain complex semantics that is not suitable to be treated as a whole in doing VideoQA. As illustrated in Figure \ref{fig:introduction}, the long-term video can be divided into four segments of different semantics and only the last segment is the key to infer the answer for the given question. Thus, naively transplanting methods for short videos to long-term VideoQA is likely to invoke redundancy and noise naturally brought by the irrelevant and less relevant parts of the video. In other words, if those parts could be trimmed off at first, the difficulty of model making the answer could be eased to an extent. 

A naturally idea for collecting relevant visual context in long-term video is to use cross-attention among video frames and question. In this way, weights in termporal series can be obtained and thus can be regared as a soft localization for reweighting frame level features. Indeed, such attempts have been made in several previous works \cite{gao2019structured, jiang2020divide}. However, the temporal attention based paradigm has its inherent defects. First, it is difficult to design an effective constraint or supervision signal for temporal attention, leading the learning of localization depending only on the final answer as supervision. Second, the peak of temporal attention distribution often appears in discontinuous frames, which cannot segment a continuous semantic activity. This makes the model lack of interpretability and affects the diagnosis of it.

To overcome the above mentioned problems, we propose a new paradigm ``locate before answering'', \textit{i.e.}, given a long-term video $V$ and a question $Q$, we first localize a relevant segment $V_{st:ed}$ which is considered sufficient to answer the question, then make the answer with only $V_{st:ed}$ and $Q$. Compared with the temporal attention based ``soft'' localization, our proposed method adopt a ``hard'' localization strategy. This is feasible and natural because human often first have a rough idea of where to find the answer, then carefully watch a specific segment of interest to find out the answer. Moreover, the pipeline becomes more interpretable since during evaluation we can also evaluate the relevance of the located segment to the question. We can also diagnose the model effectively by visualizing the question localization outputs.

To accommodate the fact that there is no temporal label of the question localization of the video in widely used VideoQA datasets \cite{xiao2021next, yu2019activitynet}, one could employ the advance in weakly supervised Video Temporal Grounding (VideoTG) \cite{anne2017localizing, gao2017tall, wang2021weakly} which specifically targets localizing a text query to a video without using explicit temporal annotation. Under this scheme, a default design could be a two-stage pipeline consisting of a question localization (QL) module and an answer prediction (AP) module, and the two modules are trained successively. However, unlike weakly supervised VideoTG in which there is no label at all, we would like to make the maximum use of the available answer annotation and let it not only serves as the supervision signal of the answer prediction module, but also explicitly guides the learning of the question localization. Therefore, we propose a novel approach named as LocAns, in which QL and AP are embedded into an end-to-end model. LocAns generates pesudo temporal label from the answer annotation on-the-fly in the training process. To be specific, we pre-define a series of proposals $V_{st:ed}$. In each training iteration, we let AP select a best proposal based on the predictive performance and use the selected proposal as the pesudo label for QL learning. To this end, there exists a mutual dependency between QL and AP in the forward pass: AP takes the outcome of QL as input to make the answer, while it also reciprocally provides the supervision signal of QL. In this way, we can supervise the learning of QL in an explicit form, which is difficult to implement in the temporal attention based paradigm. However, the dependence between AP and QL also results in a difficulty for the model converge in training. To best mitigate this effect, we train QL and AP alternatively per epoch. This strategy also frees the model from the burden of selecting the balance factor of the QL loss and AP loss which happens when the two modules are trained simultaneously.

In summary, our main contributions as follows:
\begin{itemize}
    \item We propose a new paradigm ``locate before answering'' for long-term VideoQA, which helps on removing the noise and redundancy caused by irrelevant parts of video and adds model interpretability by having the located segment as a side output.
    \item We propose a new approach named as LocAns, which embeds the question localization module and the answer prediction module into an end-to-end model, and adopt an alternatively training strategy with only the answer labels for the two modules learning.
    \item LocAns achieves state-of-the-art performance on three long-term video QA datasets NExT-QA \cite{xiao2021next}, ActivityNet-QA \cite{yu2019activitynet}, AGQA \cite{grunde2021agqa}, and our qualitative evaluation shows the effectiveness of the localization.
\end{itemize}

\label{Introduction}

\begin{figure*}[t]
\centering
\includegraphics[width=\linewidth]{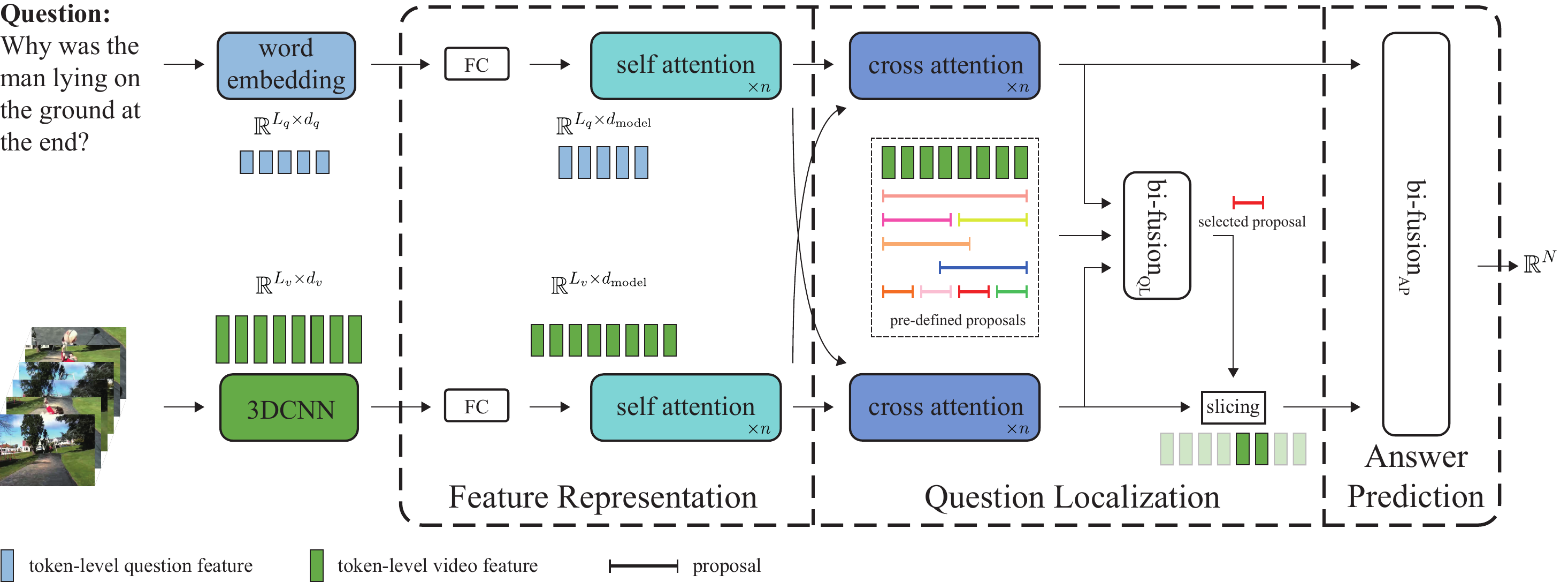}
\caption{The overall structure of the inference forward of LocAns. The proposed LocAns mainly consists of three components: a feature representation module, a question locator and a answer predictor.}
\label{fig:framework}
\end{figure*}

\section{Related Works}

\subsection{Video Question Answering}
Video Question Answering (VideoQA) \cite{jang2017tgif, zhu2017uncovering} amis to answer the nature language question in a video input, it is a natural extension of image-based VQA \cite{malinowski2014multi, antol2015vqa, yu2018beyond, yu2019deep} in video domain.
Established VideoQA works usually first extract video features using 2D CNNs followed by RNNs or 3D CNNs and language feature using GloVe \cite{pennington2014glove} or BERT \cite{devlin2018bert}, then apply a cross-modal interaction module for answer reasoning. Differing to ImageQA methods, works in VideoQA pay more attention on the context modeling between the question and video frame cross temporal series. On this basis, a few temporal attention based methods \cite{gao2019structured, jiang2020divide} conducted the initial exploration of applying question localization in VideoQA, such as Jiang \emph{et al.} \cite{jiang2020divide} proposed a question-guided spatial-temporal contextual attention network. Dang \emph{et al.} \cite{dang2021hierarchical} proposed a hierarchical object-oriented spatio-temporal reasoning networks to establish dynamic interaction between objects along video sequences. Seo \emph{et al.} \cite{seo2021attend} proposed motion-appearance synergistic network, which selectively utilizes information from two modalities based on the intention of the question.

However, most of these works were tested on early datasets such as MSVD-QA \cite{xu2017video}, MSRVTT-QA \cite{xu2017video} and TGIF-QA \cite{jang2017tgif}, in which the videos are relatively short (within 15 seconds). In recent years, several datasets have emerged for VideoQA with untrimmed long-term videos, such as NExT-QA \cite{xiao2021next}, ActivityNet-QA \cite{yu2019activitynet}, AGQA \cite{grunde2021agqa}, and TVQA \cite{lei2018tvqa}. In addition to videos and annotated question-answer pairs, TVQA is a tri-modal dataset containing subtitles, which makes this dataset more focused on language interaction between questions and subtitles. This is beyond the scope of this study. Therefore, our experiments are conducted on the rest three datasets, \textit{i.e.}, NExT-QA, ActivityNet-QA, and AGQA.

\subsection{Video Temporal Grounding}
Video Temporal Grounding (VideoTG) \cite{anne2017localizing, gao2017tall, xu2022point} aims to localize a text query to a segment in a video. Most works follow a two-stage pipeline that first generate proposals with sliding-window or other approaches, then calculate the matching score between the language query and proposals.
Weakly supervised VideoTG is a setting where no temporal label is available in order to save the expensive temporal annotation cost. Under this setting, most works \cite{mithun2019weakly, ma2020vlanet, huang2021cross} adopt the multi-instance learning (MIL) strategy to learn video-text alignment through the comparison between positive and negative samples. Mithun \emph{et al.} proposed a pioneer work named Text-Guided Attention (TGA) under the weakly supervised setting, which extracts text-guided video feature and distinguishes positive and negative moment-query pairs depending on ranking loss. Huang \emph{et al.} \cite{huang2021cross} presented Cross-sentence Mining (CRM) to explore the temporal information modeling in MIL using combinational associations among sentences.

Similar to weakly supervised VideoTG, our work is committed to locating the question to the video without explicit temporal annotation. The difference is that we leverage the guidance of question answering annotations rather than adopting MIL strategy.

\section{Methodology}
Given an untrimmed video $V$, a question $Q$ in natural language, VideoQA is to select a candidate $A$ from its searching space $\{A_i\}_{i=1}^{K}$ that best answers the question. For long-term video which often contains complex semantics, we argue that a question is likely to be related to only part of the video. To make use of this nature, we propose LocAns, an end-to-end model that first localizes the question to a certain segment in the video then predicts the answer based on only the retrieved segment. 

The overall framework is illustrated in Figure \ref{fig:framework}, the raw video and question are first individually fed into two pre-trained models to extract first-level features. Then our feature representation module encodes the temporal relation of the two modalities individually via self-attention \cite{vaswani2017attention}. The following question localization module, which employs cross-modal attention and bi-linear fusion \cite{ben2019block}, fuses the two modalities and performs proposal-based question localization. Finally the answer prediction module uses the clipped fused feature to predict the answer. 

In training the network, we are not able to directly supervise the learning of the question localizer due to the lack of existing temporal annotation. To accommodate this, we compromise to model-generated pesudo temporal annotation. In the rest of this section, the components of our network are detailed in Section \ref{sec:feature_representation_module} to \ref{sec:answer_predictor_module}, and the training strategy is explained in Section \ref{sec:cross_training}.

\subsection{Feature Representation Module}
\label{sec:feature_representation_module}
After the initial processing using pre-trained models, the video and question features $\mathbf{v}\in \mathbb{R}^{L_v\times d_v}$ and $\mathbf{q}\in \mathbb{R}^{L_q\times d_q}$ are obtained. We apply two fully connected layers to align the features to the same dimension $d_\text{model}$. To encode the token-wise sequential relations, we adopt attention transformation \cite{vaswani2017attention} globally in our model, defined as 

\begin{align}
    \text{Attn}(\mathbf{s}_q,\mathbf{s}_k,\mathbf{s}_v) = \text{softmax}(\frac{Q(\mathbf{s}_q)K(\mathbf{s}_k)^T}{\sqrt{d_\text{model}/h}})V(\mathbf{s}_v), 
\end{align}
where $\mathbf{s}_q\in\mathbb{R}^{N_q \times d_\text{model}}$, $\mathbf{s}_k,\mathbf{s}_v\in\mathbb{R}^{N_{kv} \times d_\text{model}}$, the $Q(\cdot)$, $K(\cdot)$ and $V(\cdot)$ are three independent linear transformations and $h$ denotes the number of heads. Our feature representation module applies self-attention on the visual and textual modalities individually, which produces

\begin{align}
    \mathbf{q}_\text{self}=\text{Attn}(\mathbf{q},\mathbf{q},\mathbf{q}) \in \mathbb{R}^{L_q\times d_\text{model}}, \\
    \mathbf{v}_\text{self}=\text{Attn}(\mathbf{v},\mathbf{v},\mathbf{v}) \in \mathbb{R}^{L_v\times d_\text{model}}.
\end{align}

\subsection{Question Localization (QL) Module}
\label{sec:question_locator_module}
The QL retrieves a segment of the video $V_\text{st:ed}$ such that the segment is most relevant to $Q$. To achieve this, LocAns generates $N$ total segment proposals by first defining a series of anchors with different scales, and then applying non-overlapped sliding window on the video using these anchors. To be specific, a total of 15 proposals are generated by choosing $\{1/1, 1/2, 1/3,1/4,1/5\}$ as the anchor scales in our default setting\footnote{The start and end frame are rounded down and up respectively when the number of frames cannot be divided.}. To this end, the localization is essentially transformed into a $N$-class classification problem. Since achieving reliable localization requires an understanding of the visual-textual context, we apply cross-modal attention on the self-encoded features produced by feature representation module. In this way, the cross encoding of the two modalities is respectively given by

\begin{align}
    \mathbf{q}_\text{cross}&=\text{Attn}(\mathbf{q}_\text{self},\mathbf{v}_\text{self},\mathbf{v}_\text{self}) \in \mathbb{R}^{L_q\times d_\text{model}}, \\
    \mathbf{v}_\text{cross}&=\text{Attn}(\mathbf{v}_\text{self},\mathbf{q}_\text{self},\mathbf{q}_\text{self}) \in \mathbb{R}^{L_v\times d_\text{model}}.
\end{align}

To perform the localization, we first obtain the overall feature of the sequence by max pooling $\mathbf{q}_\text{cross}$ and $\mathbf{v}_\text{cross}$ to the joint space $\mathbb{R}^{d_\text{model}}$, and subsequently applying bi-linear fusion \cite{ben2019block} to fuse the features to the proposal scores space $\mathbb{R}^N$, given by

\begin{align}
    \text{score}_\text{QL} = \text{bi-fusion}(\text{max}(\mathbf{q}_\text{cross}),\text{max}(\mathbf{v}_\text{cross})) \in \mathbb{R}^N.
\end{align}

Thus, the localization result is given by the proposal 
which gets the max of $\text{score}_\text{QL}$.

\subsection{Answer Prediction (AP) Module}
\label{sec:answer_predictor_module}
The AP makes the final prediction $A$ using only the localized segment of the video from the question localization module. To achieve this, we rule out the irrelevant part of the video by slicing the video feature $\mathbf{v}_\text{cross}$ to $\mathbf{v}_{\text{cross}_\text{st:ed}}$ where $st$ and $ed$ are the start and end frame index of the proposal selected by our question localization module. The question answering scores are generated similarly to the proposal scores, given by

\begin{align}
    \text{score}_\text{AP} = \text{bi-fusion}(\text{max}(\mathbf{q}_\text{cross}),\text{max}(\mathbf{v}_{\text{cross}_\text{st:ed}})) \in \mathbb{R}^K.
\end{align}

Thus, the answer is given by the candidate which gets the max of $\text{score}_\text{AP}$. By supervising the scores with the correct answer label, we can immediately have an answer prediction loss

\begin{align}
    \mathcal{L}_\text{AP}=\text{CrossEntropy}(\text{score}_\text{AP}, \mathbf{y}_\text{AP}),
\end{align}
where $\mathbf{y}_\text{AP}\in \mathbb{R}^K$ is the one-hot encoding of the correct answer.

\subsection{Training Strategy
\label{sec:cross_training}}

\begin{figure}
    \centering
    \includegraphics[width=\linewidth]{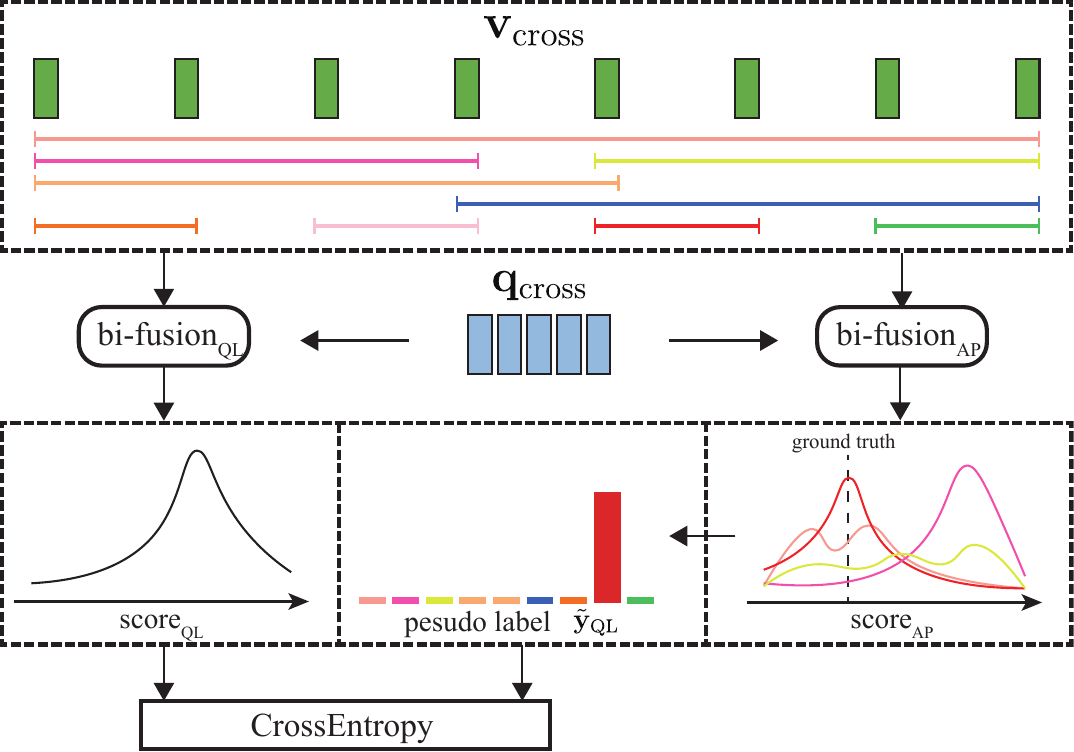}
    \caption{The generation of the pesudo label $\tilde{\mathbf{y}}_\text{QL}$ and question localization loss $\mathcal{L}_\text{QL}$.}
    \label{fig:train}
\end{figure}

\begin{algorithm}[t]
\caption{Decoupled Alternative Training in a PyTorch-like style}
\label{alg:DA}
\definecolor{codeblue}{rgb}{0.25,0.5,0.5}
\lstset{
  backgroundcolor=\color{white},
  basicstyle=\ttfamily\selectfont,
  columns=fullflexible,
  breaklines=true,
  captionpos=b,
  commentstyle=\color{codeblue},
}
\begin{lstlisting}[language=python]
# model: torch.nn.Module
# loader: torch.utils.data.DataLoader
# TOTAL_EP: number of total epochs
optimizer = Adam(model.parameters())

def odd_epoch():
    # train all modules except QL
    model.requires_grad = True
    model.QL.requires_grad = False
    for batch in loader:
        loss_ap, _ = model.forward(batch)
        loss_ap.backward()
        optimizer.step()

def even_epoch():
    # train QL only
    model.requires_grad = False
    model.QL.requires_grad = True
    for batch in loader:
        _, loss_ql = model.forward(batch)
        loss_ql.backward()
        optimizer.step()

for i in range(1, TOTAL_EP+1):
    if i % 2 == 1:
        odd_epoch()
    else:
        even_epoch()
\end{lstlisting}


\end{algorithm}

\textbf{Question Localization Loss:}
Our framework consists of two important modules that play clearly different roles, question localization module and answer prediction module. The quality of the question localization essentially determines the performance of the answer prediction: if the question localization module makes a wrong decision, theoretically the predicted answer can be no better than a random guess. Thus, instead of naively training the whole framework using $\mathcal{L}_\text{AP}$, we would like to introduce another loss term to particularly supervise the learning of the question localization. However, since there is no true label of the correct proposal nor the $st$ and $ed$ ground truth, we compromise to using pesudo label. As illustrated in Figure \ref{fig:train}, we forward pass our model using all the N proposals, and the proposal $p$ which gives the highest score on the correct answer is reciprocally used as the supervision signal that guides the learning of the question localization. In this way, the question localization loss is given by

\begin{align}
    \label{eq:loss_ql}
    \mathcal{L}_\text{QL}=\text{CrossEntropy}(\text{score}_\text{QL}, \tilde{\mathbf{y}}_\text{QL}),
\end{align}
where $\tilde{\mathbf{y}}_\text{QL}\in \mathbb{R}^N$ is the one-hot encoding of $p$.

\begin{table*}[]
\centering
\caption{Performance comparison with various existing methods on NExT-QA validation set. AccC, AccT, AccD denote the accuracy of causal, temporal, descriptive questions respectively.}
\begin{tabular}{c|ccc|ccc|cccc|c}
\hline
\multirow{2}{*}{Methods} & \multicolumn{3}{c|}{AccC} & \multicolumn{3}{c|}{AccT} & \multicolumn{4}{c|}{AccD} & \multirow{2}{*}{Acc} \\ \cline{2-11}
           & Why   & How   & All   & B\&A & Pre & All   & Cnt & Loc & Other & All   &       \\ \hline
E-VQA     & 42.31 & 42.90 & 42.46 & 46.68      & 45.85   & 46.34 & \underline{44.07} & 46.44    & 46.23 & 45.82 & 44.24 \\
ST-VQA     & 45.37 & 43.05 & 44.76 & 47.52      & \underline{51.73}   & 49.26 & 43.50 & 65.42    & 53.77 & 55.86 & 47.94 \\
Co-Mem     & 46.15 & 42.61 & 45.22 & 48.16      & 50.38   & 49.07 & 41.81 & 67.12    & 51.80 & 55.34 & 48.04 \\
HCRN       & \underline{46.99} & 42.90 & 45.91 & 48.16      & 50.83   & 49.26 & 40.68 & 65.42    & 49.84 & 53.67 & 48.20 \\
HME        & 46.52 & \underline{45.24} & 46.18 & 47.52      & 49.17   & 48.20 & 45.20 & \underline{73.56}    & 51.15 & 58.30 & 48.72 \\
HGA        & \underline{46.99} & 44.22 & \underline{46.26} & \underline{49.53} & \textbf{52.49}   & \underline{50.74} & \underline{44.07} & 72.54    & \underline{55.41} & \underline{59.33} & \underline{49.74} \\ \hline
Ours & \textbf{49.90} & \textbf{45.68}   & \textbf{48.79} & \textbf{51.21} &    50.68     &   \textbf{50.99}    &   \textbf{44.63} &   \textbf{75.93}       &    \textbf{55.74}   &    \textbf{60.88}   & \textbf{51.38}      \\ \hline
\end{tabular}
\label{tab:Next_SOTA}
\end{table*}

\begin{table}[]
\centering
\caption{Performance comparison with various existing methods on ActivityNet-QA test set.}
\begin{tabular}{c|cc}
\hline
Methods & Vid Feats            & Accuracy \\ \hline
E-VQA   & VGG + C3D            & 25.1     \\
E-MN    & VGG + C3D            & 27.1     \\
E-SA    & VGG + C3D            & 31.8     \\
MAR-VQA & VGG + C3D + Audio    & 34.6     \\
CAN     & VGG + C3D            & 35.4     \\ \hline
HGA \dag    & C3D                  & 34.6     \\ 
Ours    & C3D                  & \textbf{36.1} \\ \hline
\end{tabular}
\label{tab:Anet_SOTA}
\end{table}

\begin{table}[]
\centering
\caption{Performance comparison with various existing methods on AGQA v2 test set.}
\begin{tabular}{c|cc}
\hline
Methods & Vid Feats            & Accuracy \\ \hline
PSAC   & ResNet & 40.18 \\
HME    & ResNet + ResNeXt & 39.89 \\
HCRN    & ResNet + ResNeXt & 42.11 \\ \hline
Ours    & ResNet & \textbf{47.07} \\ \hline
\end{tabular}
\label{tab:AGQA_SOTA}
\end{table}

\textbf{Decoupled Alternative (DA) Training:}
With $\mathcal{L}_\text{QL}$ as an augmentation of the default  $\mathcal{L}_\text{AP}$, the naive way of training the network is to sum the two loss terms with a weight factor $\lambda$ such that

\begin{align}
    \label{eq:total_loss}
    \mathcal{L}=\mathcal{L}_\text{AP}+\lambda \mathcal{L}_\text{QL}.
\end{align}

However, as also can be seen in our ablation study in Section \ref{sec:abl}, finding the optimum $\lambda$ is difficult since the optimum $\lambda$ varies across different datasets, and also costly since the tuning needs many rounds of experiments on the training set. 

To avoid the above disadvantages, we design a DA training strategy which does not rely on weighting the two loss terms with $\lambda$. Observing the design of our model, the quality of QL determines the lower bound of AP's performance, while AP reciprocally provides the supversion signal to QL. Inspired by the idea of Expectation Maximization (EM) algorithm \cite{dempster1977maximum}, we make use of this mutual effect and decouple the naive training strategy that updates the two correlated module QL and AP jointly using the bundled loss in Equation \ref{eq:total_loss} into two phases. As illustrated by the pesudo codes in Algorithm \ref{alg:DA}, phase 1 trains the whole model except for QL using $\mathcal{L}_\text{AP}$ while phase 2 tunes only the parameters of QL supervised by $\mathcal{L}_\text{QL}$. Thus, there is no need for setting $\lambda$. We alternate the two phases per epoch to ensure a sufficient and balanced learning dynamic.

\section{Experiments}

\subsection{Datasets}
Since we focus on the scenario of long-term VideoQA, we conduct our experiments on the following three modern VideoQA datasets containing videos longer than most counterparts.
\paragraph{NExT-QA} Videos in NExT-QA \cite{xiao2021next} are sourced from YFCC-100M \cite{thomee2016yfcc100m}, and then manually annotated. The NExT-QA dataset contains 5,440 videos with average duration of 44 seconds, and 52,044 question-answer pairs in total. In this dataset, each question is given with 5 answer candidates and 1 of them is correct.

\paragraph{ActivityNet-QA} ActivityNet-QA \cite{yu2019activitynet} is an automatically re-annotated version of ActivityNet \cite{caba2015activitynet} based on existing video descriptions. It contains 5,800 videos with average duration of 180 seconds, and 58,000 question-answer pairs (32k/18k/8k for train/val/test split). In this dataset, only a correct answer is given with no other candidates, hence we select the top 1,000 most common answers in the training set as the answer vocabulary. This rules out 4,883 question-answer pairs from the training set.

\paragraph{AGQA} AGQA \cite{grunde2021agqa} is a novel video question answering benchmark designed for evaluating models' reasoning abilities concerning compositional spatiot-emporal interactions. We use AGQA v2, which consists of 9.7K videos and 2.27 million QA pairs, with an average video duration of 30 seconds.

\subsection{Evaluation Metric}
Our method is evaluated by the accuracy of the question answering. Since the temporal annotation is not available, we evaluate our question locating performance with human evaluated qualitative examples.

\subsection{Implementation Details}
The visual feature dimensions of NExT-QA (Res + I3D), ActivityNet-QA (C3D), and AGQA (ResNet) are 4096, 500, and 2048, respectively. The textual feature dimensions of NExT-QA (BERT), ActivityNet-QA (6B GloVe), and AGQA (6B GloVe) are 768, 300, and 300, respectively. For the model hyperparameters, we adopt 2-layer self attention and 1-layer cross attention, and the overall model dimension $d_\text{model}$ on NExT-QA, ActivityNet-QA, AGQA is respectively set to 1024, 512, and 512. 
In the details of model implementation, we use the standard 8 heads for each encoding module in the uni-modal encoding and the cross-modal encoding. Each module is subsequently followed by a two-layer feed-forward module activated by ReLU \cite{nair2010rectified} to further enhance the encoding capacity. Moreover, we follow the standard configuration of multihead attention modules, where layernorm \cite{ba2016layer}, dropout \cite{srivastava2014dropout}, position embedding \cite{devlin2018bert} and residual connection \cite{he2016deep} are applied. In terms of training, we optimize our model using the Adam optimizer \cite{kingma2014adam} with learning rate of $5\times 10^{-5}$ for NExt-QA and $1\times 10^{-4}$ half decaying on plateau for ActivityNet-QA and AGQA. All experiments are conducted on a Nvidia Tesla V100 GPU with 32GB memory in the condition of 64 batch size.

\begin{table}[t]
\centering
\caption{Module ablation on NExT-QA dataset.}
\begin{tabular}{l|c}
\hline
Variants               & Accuracy \\ \hline
w/o QL & 48.32    \\
w/o $\mathcal{L}_\text{QL}$        & 50.05    \\ 
w/ Soft Attention QL & 49.37  \\ \hline
Full Model             & 51.38     \\ \hline
\end{tabular}
\label{tab:ablation}
\end{table}

\begin{table}[t]
\centering
\caption{Training strategy comparison on NExT-QA dataset.}
\begin{tabular}{c|cc}
\hline
Training Strategy        & Con. Epoch & Accuracy \\ \hline
Bundled Training        & 43         & 51.50  \\
DA training              & 12         & 51.38  \\ \hline
\end{tabular}
\label{tab:training}
\end{table}

\subsection{Performance Comparison}
To demonstrate the effectiveness of our proposed LocAns, we compare it with several classic and current state-of-the-art VideoQA methods: E-VQA \cite{antol2015vqa}, E-MN \cite{sukhbaatar2015end}, E-SA \cite{yao2015describing}, ST-VQA \cite{jang2017tgif}, Co-Mem \cite{gao2018motion}, HCRN \cite{le2020hierarchical}, HME \cite{fan2019heterogeneous}, HGA \cite{jiang2020reasoning}, MAR-VQA \cite{zhuang2020multichannel}, CAN \cite{yu2019compositional}. Table \ref{tab:Next_SOTA} to Table \ref{tab:AGQA_SOTA} summarize the performance comparison on NExT-QA, ActivityNet-QA, and AGQA, respectively.
\paragraph{NExT-QA} For a fair comparison, the compared methods on NExT-QA dataset use the same video and language features. The videos are uniformly sampled into 16 clips and each clip contains 16 consecutive frames. Each clip feature is composed of appearance feature extracted from ResNet-101 \cite{he2016deep} pre-trained on ImageNet \cite{deng2009imagenet} and motion feature extracted from 3D ResNeXt-101 \cite{xie2017aggregated} pre-trained on Kinetics \cite{kay2017kinetics}. The questions and candidate answers use pre-trained BERT \cite{devlin2018bert} for sentence embedding.

For the results in Table \ref{tab:Next_SOTA}, we have the following observations: 1) Our proposed LocAns outperforms other methods regarding the overall accuracy. Compared with the previous best method HGA, we have a 1.64\% improvement. 2) Our method also achieves promising performance on most fine-grained categories, suggesting that ``locate before answering'' is widely applicable without significant limitation on the question type. 3) The most significant accuracy increase of our method on AccC suggests a particular necessity of localization in long-term video for answering complex causal reasoning questions.

\paragraph{ActivityNet-QA} For ActivityNet-QA dataset, the existing methods are not unified in video features. We choose the methods with widely used VGG-16 \cite{simonyan2014very} and C3D \cite{tran2015learning} features for comparison. Other methods using S3D \cite{xie2018rethinking} or object detection \cite{ren2015faster} features such as \cite{yu2020long}, and methods using additional training data such as \cite{seo2021look, yang2021just}, are not within the scope of our comparison. On the language side, each word token is represented by the pre-trained 300-dimensional GloVe embedding \cite{pennington2014glove}. As we can see in Table \ref{tab:Anet_SOTA}, our proposed LocAns reaches the highest accuracy even we only use the C3D feature. We have a 1.5\% improvement over HGA that uses the same C3D features. Compared with MAR-VQA with additional VGG and audio features, we also have a 1.5\% improvement.

\paragraph{AGQA} For AGQA dataset, our proposed LocAns use the appearance feature extracted by pre-trained ResNet-101 as other methods. As shown in Table \ref{tab:AGQA_SOTA}, LocAns significantly outperforms other methods when using only ResNet features. On AGQA, LocAns achieves an accuracy of 47.07\%, which is nearly 5\% higher than the previous best-performing method, HCRN.

\subsection{Ablation Studies}
\label{sec:abl}
\begin{figure}[t]
    \centering
    \subfigure[NExT-QA]{\includegraphics[width=\linewidth]{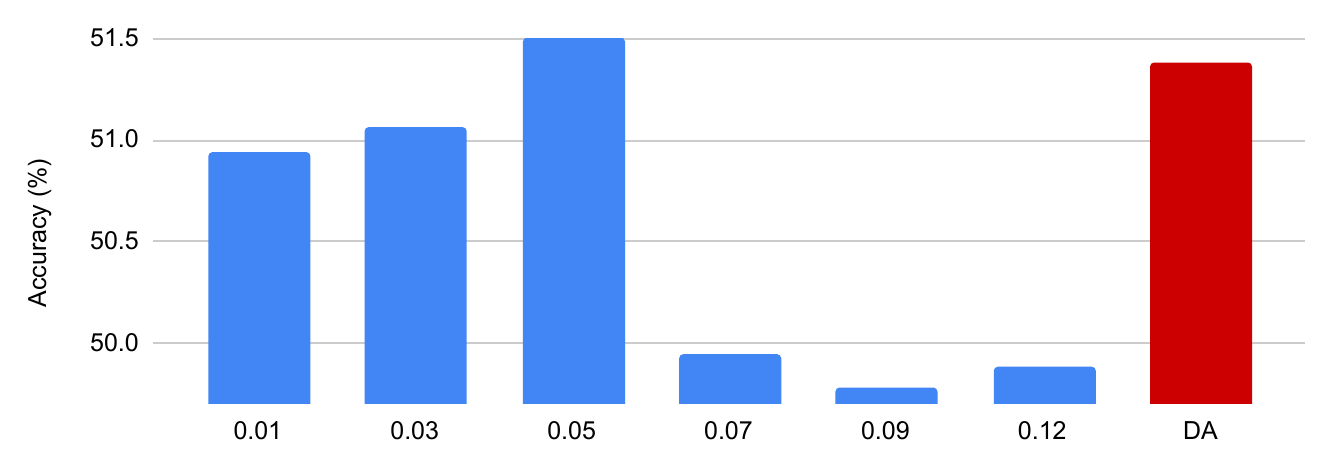}}
    \quad
    \subfigure[ActivityNet-QA]{\includegraphics[width=\linewidth]{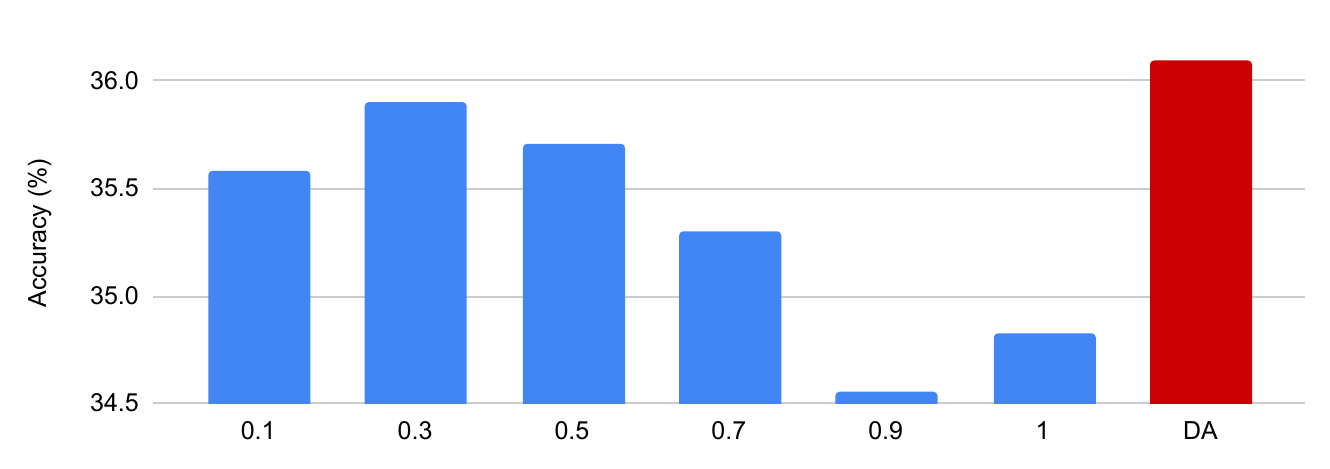}}
    
    \caption{Performance comparison with bundled training with different $\lambda$ and proposed DA training on two datasets.}
    \label{fig:hyperparameter}
\end{figure}

\begin{table}[t]
\centering
\caption{Influence of pre-defined proposals on NExT-QA dataset.}
\begin{tabular}{c|cc}
\hline
Scale                        & Total Pro. & Accuracy \\ \hline
\{1/3,  1/2, 1\}             & 6      & 51.07    \\
\{1/7,  1/8\}                & 15     & 50.46    \\
\{1/10,  1/5, 1/3,  1/2, 1\} & 21     & 51.24\\ \hline
\{1/5, 1/4, 1/3, 1/2, 1\}    & 15     & 51.38    
\\ \hline
\end{tabular}
\label{tab:proposal}
\end{table}

In order to comprehensively analyze our proposed LocAns, we conduct extensive and in-depth ablation experiments on NExT-QA validation split.

\paragraph{Effectiveness of Components} We first design three model variants to verify the effectiveness of the key modules in LocAns. The results are listed in Table \ref{tab:ablation}. In variant 1 (w/o QL), we remove the entire QL. Variant 2 (w/o $\mathcal{L}_\text{QL}$) keeps the full model structure, but only uses $\mathcal{L}_\text{AP}$ for training. In order to verify the effectiveness of our hard localization with proposals, we also implement a soft localization method variant 3 (w/ Soft Attention QL) for comparison. It takes the question-to-video attention score as the soft localization result and weighted average the video features in the temporal dimension 
instead of a hard clipping.

\begin{figure}[ht]
    \centering
    \includegraphics[width=0.9\linewidth]{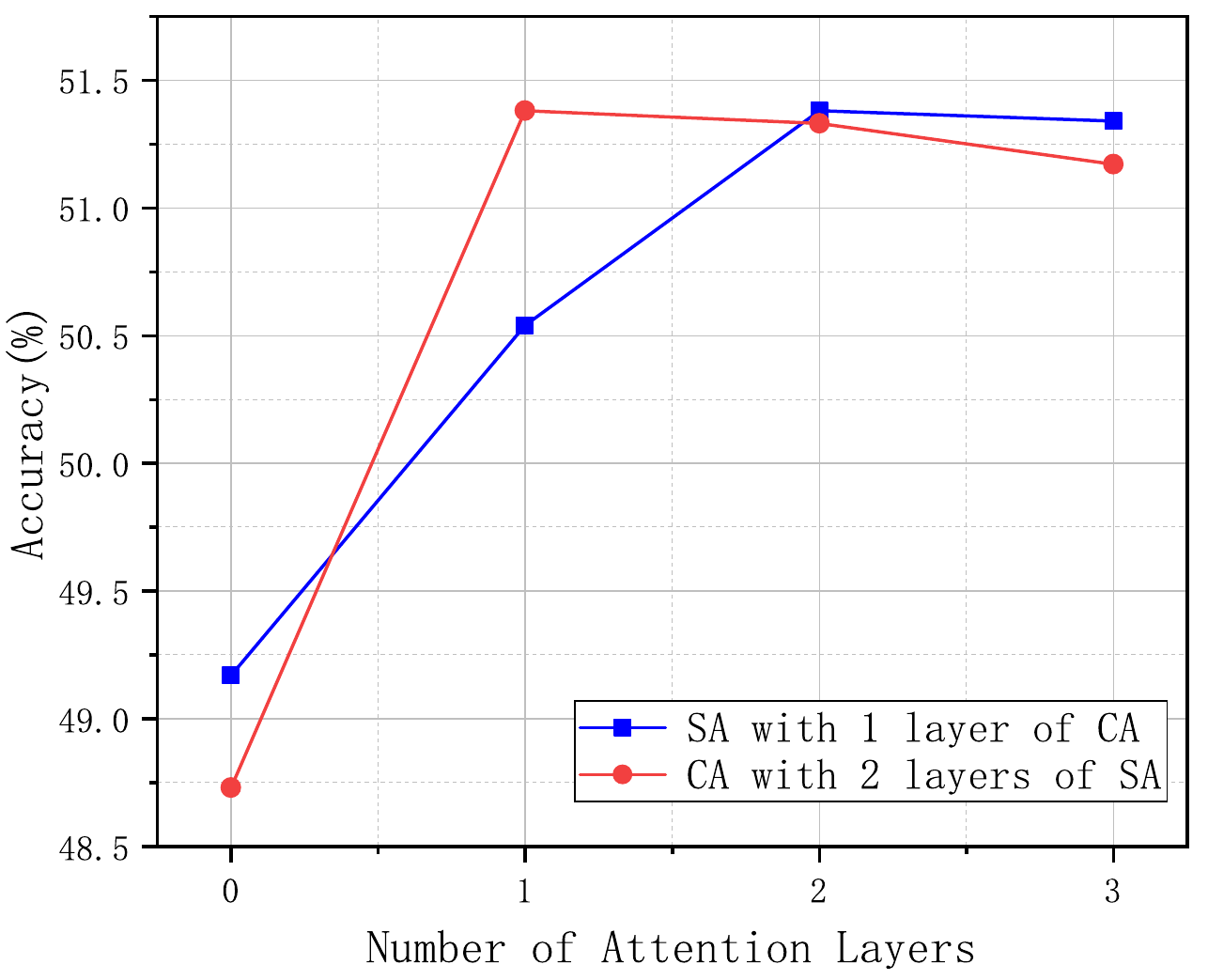}
    \caption{Performance of LocAns on NExT-QA with respect to different layer numbers of self-attention and cross-attention. SA and CA denote self-attention and cross-attention respectively.}
    \label{fig:layer}
\end{figure}

According to the results in Table \ref{tab:ablation}, we have the following observations: First, our full model obtains a significant improvement over variant 1 and variant 2 by 3.06\% and 1.33\%, respectively. Variant 2 suggests that the localization ability learned in a weakly-supervised way, that is, using answer head to generate pseudo label of localization module, can be directly fed back to the accuracy of question answering. Second, variant 2 has an absolute improvement of 1.73\% compared with variant 1. This demonstrates that the localization module can learn the location ability to a certain extent for question answering performance improvement even without the help of pseudo label. Finally, without additional supervision, the accuracy of variant 3 is 0.68\% lower than that of variant 2, suggesting the advantage of hard proposals over soft token-level scores.

\paragraph{Necessity of DA Training} As we discussed in Section \ref{sec:cross_training}, we can naively train our model with the bundled loss in Equation \ref{eq:total_loss}. In the context of bundled training, the value of $\lambda$ determines the importance of $\mathcal{L}_\text{QL}$, thus affecting the gradient optimization direction of the whole network. However, the calculation of $\mathcal{L}_\text{QL}$ depends on the quality of the pseudo labels provided by AP, which is not trustworthy at the beginning of training. Therefore, the training may collapse from the unreliable supervision of $\mathcal{L}_\text{QL}$ when $\lambda$ is set too large. Figure \ref{fig:hyperparameter} reports the effect of hyperparameter $\lambda$ under this training strategy. It can be observed that the final performance is very sensitive with respect to $\lambda$ on both datasets. The performance tends to plummet when $\lambda$ is too large ($\lambda\geq 0.07$ on NExT-QA and $\lambda\geq 0.9$ on ActivityNet-QA), which supports our theoretical analysis. Moreover, we find that the optimum values of $\lambda$ are largely different on the two datasets, which again suggests the difficulty of tuning $\lambda$.

In contrast, our proposed DA training (red histogram in Figure \ref{fig:hyperparameter}) reaches the same level of performance as bundled training without any hyperparameter tuning. Moreover, it is worthwhile to note that DA training promotes the convergence. As shown in Table 4, DA training consumes 12 epochs for convergence, which is less than 1/3 of bundled training.

\paragraph{Influence of Pre-defined Proposals} The pre-defined proposals play a key role in our approach. Therefore, we conduct this ablation study targeting the influence of different combinations of proposals.
As we listed in Table \ref{tab:proposal}, we selected four groups of proposals for comparison, and the last group (Group 4) is our default setting. Compared with Group 4, Group 1 has fewer proposals, Group 2 has the same number of proposals but the anchor scales are different, and Group 3 has more proposals.

From the results in Table \ref{tab:proposal}, we have the following observations: 1) The performance gap between different groups is trivial, which suggests that LocAns is robust to different pre-defined proposals. 2) Too few or too many proposals will cause a certain degree of performance degradation. We speculate that this is because on the one hand, too few proposals mean coarse-grained localization, and on the other hand, too many proposals will make proposal classification difficult. 3) In the context of weakly supervised localization, coarse-grained anchor is better than too fine-grained anchor: the accuracy of Group 2 is 0.92\% lower than Group 4.

\paragraph{The number of attention layers} Figure \ref{fig:layer} summarizes the effects of the number of self-attention and cross-attention layers. From the experimental results, we can conclude that the performance will be significantly improved at the beginning and reach a bottleneck with slight drop as the number of attention layers increases. The best choice (\textit{i.e.}, our default setting) is two layers of self-attention and one layer cross-attention. Overfitting may be responsible for the performance degradation when we use deep attention layers. When the attention layer is set to 0, there is a significant decrease in performance. For instance, when the cross-attention layer is reduced from 1 to 0, the model's accuracy drops from 51.38\% to 48.73\%, demonstrating a performance gap of 2.65\%. This difference in performance reveals the crucial role of cross-attention in integrating contextual information between language and visual input.

\begin{figure*}[ht]
    \centering
    \includegraphics[width=0.83\textwidth]{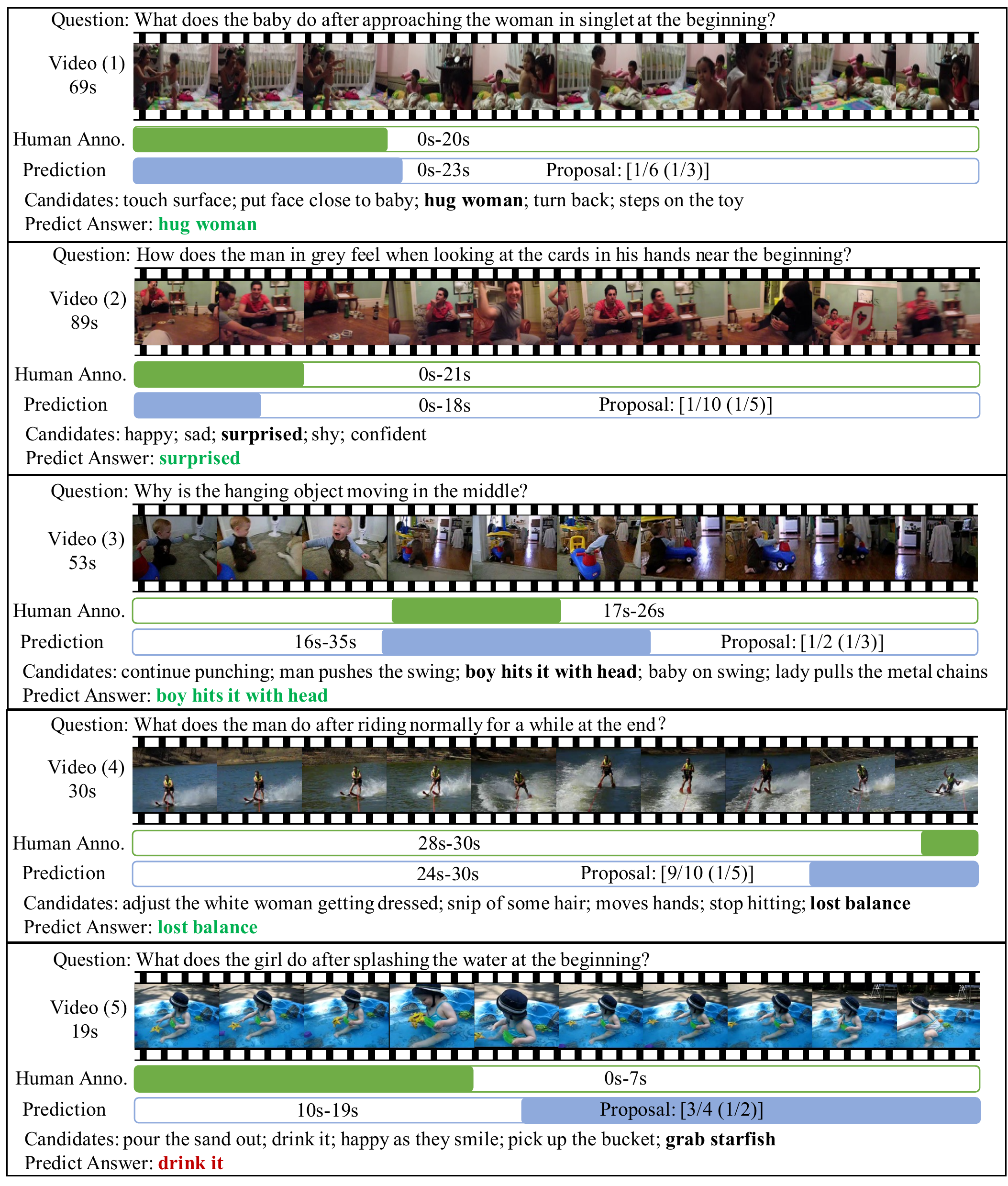}
    \caption{Some visualized examples from NExT-QA validation set. The first two are successful cases, and the third is a failing case. The green bar is the temporal boundary annotated by the author, and the blue bar is the prediction of the localization module. The two values of the proposal on the prediction bar represent the center and scale of the temporal interval it represents. For example, [1/2 (1/3)] indicates that the center is located at the 1/2 of the video, and the interval scale is 1/3 of the total video duration.}
    \label{fig:visualization}
\end{figure*}

\subsection{Qualitative Examples}
\label{sec:qual_example}
To show the reliability and interpretability of our proposed LocAns more concretely, we visualize some examples in Figure \ref{fig:visualization}. These samples can give us a rough impression of the localization performance of LocAns on the one hand, they can also provide explanations for the final answering results on the other hand. Video (1) to Video (4) are successfully answered samples, they demonstrate the effectiveness of our approach from two aspects. For Video (1) and Video (2), LocAns locates the question related segments accurately. In Video (1) the baby and the woman had an obvious hugging behavior, thus trimming off the irrelevant visual contents for around 50 seconds. For Video (3) and Video (4), even if the localization does not match the human annotated boundary perfectly (the prediction interval is slightly wider than the human annotated interval), it still contributes to the correct answer prediction. Inevitably, there are also failure cases like Video (5), where the key frame of the girl grasping starfish and the localized segment are staggered completely. We consider that this may be caused by the visual difference between frames in Video (5) is very small. The entire video is about a girl sitting in the blue water, which makes fine-grained action (grab starfish) detection difficult. 

From these visualized examples, we have the following observations: 1) Most of the successful cases have precise temporal prediction, while the failure cases are the opposite. 2) Although the localization is not completely precise in some cases (\textit{e.g.}, Video (3) and Video (4)), it still filters out part of noise and redundancy. 3) In some extreme cases, the temporal prediction will be completely wrong, which will cause the input of answer predictor being untrustworthy. This further illustrates the importance of question localization in long-term videos.


\section{Limitations}
The main limitation of this study is that there is no ground truth of the temporal label. One consequence of this fact is that we are only able to generate pesudo label and perform proposal based localization. Thus the granularity and the accuracy of localization are largely limited. Another consequence lead by the absence of temporal annotation is that we can only evaluate our question localization module with less-indicative qualitative examples instead of overall quantitative scores. In the future, a VideoQA dataset containing annotated temporal boundary of the question would largely boost the study in this direction.

\section{Conclusion}
In this paper, we explore a new paradigm of first locating the question and then making the answer for long duration VideoQA to tackle the difficulty brought by more complex visual semantics. Based on this paradigm, we propose LocAns, a novel algorithm which integrates a question localization module and a answer prediction module into an end-to-end model. Extensive quantitative and qualitative experiments on three long-term VideoQA datasets NExT-QA, ActivityNet-QA, and AGQA demonstrate the effectiveness of our method. Unfortunately, we can not evaluate our question localization performance quantitatively due to the absence of long-term temporal annotated VideoQA dataset, and we leave this as our future work. We hope that the ``localization first'' paradigm could be potentially generalized to other tasks related to long-term video understanding.



\ifCLASSOPTIONcaptionsoff
  \newpage
\fi

\bibliographystyle{IEEEtran}
\bibliography{IEEEfull}

\begin{IEEEbiography}[{\includegraphics[width=1in,height=1.25in,clip,keepaspectratio]{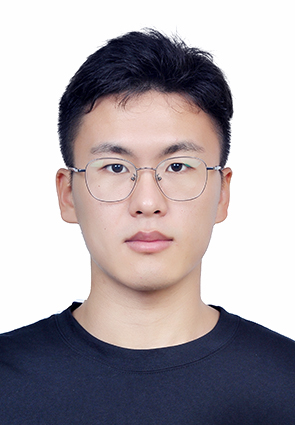}}]{Tianwen Qian} 
received his B.E. degree from Dalian University of Technology, Dalian, China, in 2019. He is currently pursuing his Ph.D. degree in Computer Science at Academy for Engineering and Technology, Fudan University. His research interest is focused on Vision-Language Understanding, especially Visual Question Answering.
\end{IEEEbiography}

\begin{IEEEbiography}[{\includegraphics[width=1in,height=1.25in,clip,keepaspectratio]{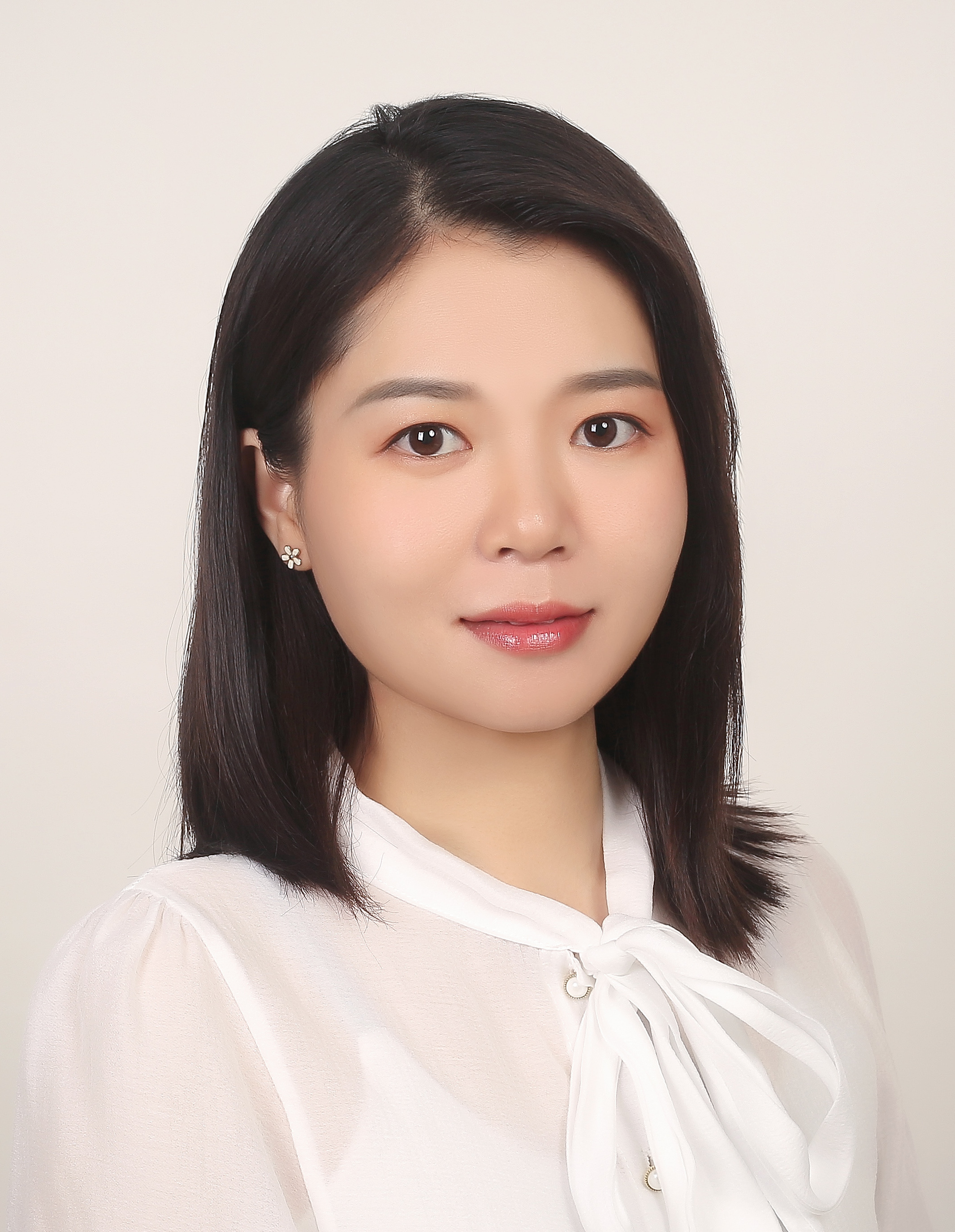}}]{Jingjing Chen} 
is now an associate professor at the School of Computer Science, Fudan University. Before that, she was a postdoc research fellow at the School of Computing in the National University of Singapore. She earned her Ph.D. degree in Computer Science from the City University of Hong Kong in 2018. Her research expertise spans robust AI and multimedia content analysis, including video content analysis, cross-modal retrieval and food recognition.
\end{IEEEbiography}

\begin{IEEEbiography}[{\includegraphics[width=1in,height=1.25in,clip,keepaspectratio]{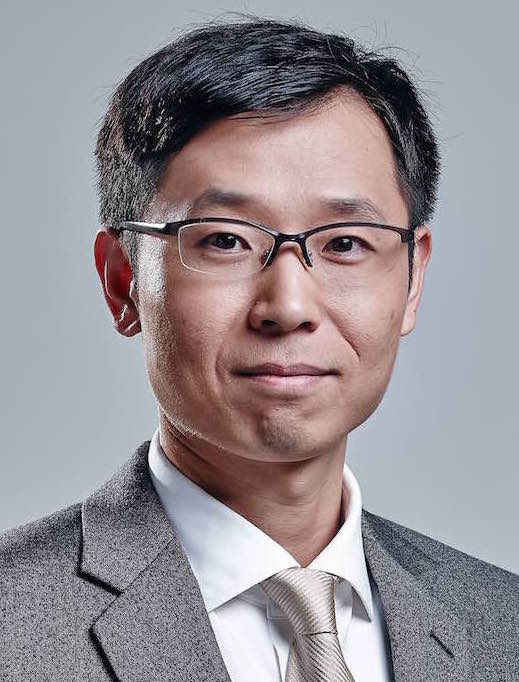}}]{Yu-Gang Jiang} received the Ph.D. degree in Computer Science from City University of Hong Kong in 2009 and worked as a Postdoctoral Research Scientist at Columbia University, New York, during 2009-2011. He is currently a Professor of Computer Science at Fudan University, Shanghai, China. His research lies in the areas of multimedia, computer vision, and robust and trustworthy AI. His work has led to many awards, including the inaugural ACM China Rising Star Award, the 2015 ACM SIGMM Rising Star Award, the Research Award for Excellent Young Scholars from NSF China, and the Chang Jiang Distinguished Professorship appointed by Ministry of Education of China.
\end{IEEEbiography}

\end{document}